\documentclass[review]{elsarticle}
\usepackage{amsmath}
\usepackage{multirow}
\usepackage{multicol}
\usepackage{graphicx}
\usepackage{color,colortbl}
\usepackage{caption}
\usepackage{makecell}
\usepackage{float}
\usepackage{array}
\usepackage{stackengine}
\usepackage{subfigure}
\usepackage{placeins}
\usepackage{indentfirst}
\usepackage[normalem]{ulem}
\usepackage{wrapfig}
\usepackage{lscape}
\usepackage{rotating}
\usepackage{epstopdf}
\usepackage{subcaption}
\usepackage{mathrsfs} 
\usepackage{amsmath} 
\usepackage{algorithm}
\usepackage{algorithmic}
\usepackage{float}
\usepackage{multirow}
\usepackage{mathtools}
\usepackage{lineno,hyperref}
\usepackage{amsfonts,amssymb}
\usepackage{booktabs}
\usepackage{endnotes}
\usepackage{multicol}
\usepackage{booktabs} 

\usepackage[utf8]{inputenc} 
\usepackage[T1]{fontenc}    
\usepackage{hyperref}       
\usepackage{url}            
\usepackage{booktabs}       
\usepackage{amsfonts}       
\usepackage{nicefrac}       
\usepackage{microtype}      
\usepackage{xcolor}         

\modulolinenumbers[5]

\journal{Journal of \LaTeX\ Templates}

\begin{document}

\begin{frontmatter}

\title{Dynamic Accumulated Attention Map for Interpreting Evolution of Decision-Making in Vision Transformer}

\author[mymainaddress]{Yi Liao}  

\author[mymainaddress]{Yongsheng Gao\corref{mycorrespondingauthor}}
\cortext[mycorrespondingauthor]{Corresponding author}
\ead{yongsheng.gao@griffith.edu.au}

\author[mysecondaryaddress,mythirdaryaddress]{Weichuan Zhang}

\address[mymainaddress]{School of Engineering and Built Environment, Griffith University, Australia}

\address[mysecondaryaddress]{Institute for Integrated and Intelligent Systems, Griffith University, Australia}

\address[mythirdaryaddress]{School of Electronic Information and Artificial Intelligence, Shaanxi University of Science and Technology, China}






\begin{abstract}
Various Vision Transformer (ViT) models have been widely used for image recognition tasks. However, existing visual explanation methods can not display the attention flow hidden inside the inner structure of ViT models, which explains how the final attention regions are formed inside a ViT for its decision-making. In this paper, a novel visual explanation approach, Dynamic Accumulated Attention Map (DAAM), is proposed to provide a tool that can visualize, for the first time, the attention flow from the top to the bottom through ViT networks. To this end, a novel decomposition module is proposed to construct and store the spatial feature information by unlocking the [class] token generated by the self-attention module of each ViT block. The module can also obtain the channel importance coefficients by decomposing the classification score for supervised ViT models. Because of the lack of classification score in self-supervised ViT models, we propose dimension-wise importance weights to compute the channel importance coefficients. Such spatial features are linearly combined with the corresponding channel importance coefficients, forming the attention map for each block. The dynamic attention flow is revealed by block-wisely accumulating each attention map. The contribution of this work focuses on visualizing the evolution dynamic of the decision-making attention for any intermediate block inside a ViT model by proposing a novel decomposition module and dimension-wise importance weights. The quantitative and qualitative analysis consistently validate the effectiveness and superior capacity of the proposed DAAM for not only interpreting ViT models with the fully-connected layers as the classifier but also self-supervised ViT models. The code is available at \url{https://github.com/ly9802/DynamicAccumulatedAttentionMap}.
\end{abstract}


\end{frontmatter}


\section{Introduction}
Vision Transformer (ViT) models have been widely applied to real-world scenarios such as autonomous driving~\cite{RangViT}, UAV-based semantic segmentation~\cite{uavformer}, medical image analysis~\cite{medvit}, and hyperspectral image processing~\cite{SMGUNet}. Meanwhile, the Transformer architecture~\cite{Transformer} has become the dominant paradigm in natural language processing (NLP),  bridging the gap between images and languages with foundational models that enable text-to-image and image-to-text generation. Prominent examples include generative systems like ChatGPT.

In domains such as medical image analysis and autonomous driving, model interpretability and trustworthy decision making are crucial. Visual explanation methods are able to uncover the specific regions a model attends to when making decisions, thereby enhancing transparency and reliability. For example, visual explanation techniques have already been incorporated to develop improved models for skin cancer classification~\cite{skincander} and brain tumor segmentation on MRI images~\cite{wsssMRI}, underscoring the practical value of model explainability.

Despite the remarkable performance of ViTs~\cite{ViT} and their variants~\cite{DeiT,T2TViT,CrossViT,DynamicViT,CaiT,SwinTransformer,LVViT,DINO,XCiT,EsViT,CLEViT,ScopeViT} in these fields, revealing the decision-making processes inside ViT models remains a significant challenge. Existing explanation methods are incapable of revealing the evolution process of the discriminative attention. The mystery hidden in their inner structure has not yet been explored in the computer vision community. The interpretability of ViT models involves the problem that what regions on images are learned by models to make a specific decision. Visualizing the decision-making regions on images will make the prediction more trustworthy and enable the neural networks more transparent. To satisfy this demand, visual explanation methods~\cite{GradCAM,GradCAM++,XGrad-CAM,LayerCAM,AblationCAM,ScoreCAM,TransformerLRP,GAE,IIA,TaylorDecomposition} are employed to generate the explanation maps that display the attention regions supporting the decision~\cite{pseudo}.      

However, these explanation methods fail to reveal the evolution process of the discriminative attention. The evolution of attention will be beneficial to knowing how ViT models learn the features of the target object. Moreover, the attention flow can qualitatively analyse the effect of the module inserted into ViT models (e.g., cross-covariance attention (XCA) module~\cite{XCiT}, LayerScale module~\cite{CaiT}, and Token-to-Token (T2T) module~\cite{T2TViT}) in terms of forming the attention of the object. It will be helpful for designing more advanced ViT models or improving the current ViT models. In addition, what contribution each intermediate ViT block makes can be known by qualitatively or quantitatively comparing the attention maps of each intermediate block, so the block making the greatest contribution can be identified with the help of attention flow. Therefore, it is necessary for a visual explanation method to interpret the intermediate layers of neural networks. 

It is well known that the [class] token is used for decision-making in ViT models for image classification. The [class] token is generated by self-attention (SA) module when implementing the operation of $\text{softmax}(\frac{QK^T}{\sqrt{d}})V$. However, the SA module does not conserve the spatial semantic information in the output token sequence during the generation process of the [class] token. Therefore, it lacks a key to unlock the black-box mechanism in the SA module to store the spatial semantic feature information. Moreover, the existing explanation methods fail to visualize the explanation for the decisions made by using k-nearest neighbour (KNN) algorithm~\cite{NPID} in self-supervised ViT models.

In this work, we propose a novel and efficient visual explanation method named Dynamic Accumulated Attention Maps (DAAM) to visualize the attention flow hidden in ViT architectures with the FC layer as the classifiers and self-supervised ViT models. Firstly, the proposed module is inserted into the SA module of each ViT block to construct and store the spatial semantic feature information by focusing on the [class] token. Secondly, the channel-wise importance coefficients are calculated by decomposing the classification score. The attention flow for each intermediate ViT block is formed by dynamically accumulating the combination of the constructed semantic feature and channel-wise importance coefficients block-wisely. For self-supervised ViT models, the channel-wise importance coefficients are calculated by decomposing the inner product between the [class] token and the dimension-wise importance weight to the decision-making similarity score. Future research will focus on leveraging DAAM’s visual explanation maps to guide the design of novel model architectures for practical applications including medical image analysis and autonomous driving. We also plan to optimize DAAM by relaxing its reliance on the [class] token, thereby enhancing its adaptability to diverse scenarios. Furthermore, integrating DAAM with iterative integration techniques may broaden its applicability to CNN-based models, ultimately expanding the scope of visual explainability across different deep learning paradigms. 

The main contributions of this work can be summarised as follows, 
\begin{itemize}
    \item A novel accumulation explanation technique is proposed, for the first time, to reveal the evolution of attention maps, i.e. an attention flow from inside (the input block, intermediate blocks) to outside (output block) through a ViT architecture when making its classification decision.
    \item We design a novel decomposition module that can preserve the spatial semantic information during the calculation of the [class] token by unlocking the self-attention mechanism.
    \item The dimension-wise importance weight is proposed to construct the explanation maps for any intermediate block inside self-supervised ViT models.
    \item The effectiveness and superiority of the proposed method are demonstrated by the qualitative visualization and quantitative experiments applied on 7 ViT models with FC layer as the classifier and 2 self-supervised ViT models.
\end{itemize}
\section{Related Work}
\label{relatedwork}
\subsection{ViT Models for Image Classification}
Transformer was initially proposed in~\cite{Transformer} for addressing the problems in the field of Natural Language Processing (NLP). Inspired by~\cite{Transformer}, researchers developed ViT~\cite{ViT} for image classification. To achieve excellent accuracy, ViT~\cite{ViT} should be pretrained well on a large-scale dataset before being transferred to tasks with small datasets due to the lack of inductive bias. To handle this problem, various training strategies~\cite{DeiT,LVViT} for ViT models are proposed. For example, DeiT~\cite{DeiT} uses an extra token for knowledge distillation. LV-ViT~\cite{LVViT} utilizes a position-specific label of each patch to implement supervised training. Mix-ViT~\cite{MixViT} mixes the high-level tokens for prediction. ReViT~\cite{ReViT} can capture and conserve discriminative low-level features by learning residual attention. ScopeViT~\cite{ScopeViT} can efficiently capture multi-scale features by utilizing multi-scale self-attention and global scale dilated attention. GasHis-Transformer~\cite{gashisTransformer} also learns multi-granularity representations via the position-encoded module and convolution module. Some studies~\cite{CaiT,T2TViT,SwinTransformer,CrossViT} aim to improve the architecture of ViT for better accuracy. CaiT~\cite{CaiT} leverages more self-attention blocks before adding the learnable class token and proposes the CaiT optimization training approach. T2T-ViT~\cite{T2TViT} adds Token-to-Token (T2T) modules to obtain the high quality of tokens that can model local structures such as edges and lines. Supervised ViT models always employ  the FC layer as the classifier to perform image classification tasks. The FC layer should be well trained under the supervision of image-level labels before testing. Compared with supervised ViT models~\cite{ViT,DeiT,T2TViT,CaiT,DynamicViT,LVViT}, self-supervised ViT models~\cite{DINO,XCiT,EsViT} are trained without the image-level label. DINO~\cite{DINO} is one of the successful self-supervised ViT models, which creatively uses self-distillation without relying on labels as the training strategy and achieves attractive accuracies. XCiT~\cite{XCiT} stacked multiple cross-covariance attention (XCA) blocks before self-attention modules and EsViT~\cite{EsViT} adopts the multi-stage architecture to improve the accuracy. The self-supervised ViT models~\cite{DINO,XCiT,EsViT} choose the memory bank~\cite{NPID} as the classifier. Visually explaining the prediction made by memory bank (KNN algorithm) for self-supervised ViT models is one of the contribution of this work. 

\subsection{Interpretability for ViT Models}
CAM-based methods~\cite{GradCAM,GradCAM++,XGrad-CAM,LayerCAM,AblationCAM,ScoreCAM} are developed to generate the explanation maps for CNN-based models that innately keep decision-making feature information (feature map). Generating visual explanation indeed rely on the spatial semantic information. However, in ViT models, the [class] token, which is used for decision-making, does not inherently store the spatial semantic information. This limitation renders CAM-based methods ineffective for ViT models. To interpret ViT architectures, several techniques have been developed. TransformerLRP~\cite{TransformerLRP} employs Layer-wise Relevance Propagation to compute local relevance scores for each attention head, and then propagates these scores through both attention layers and skip connections. Focusing on bi-modal interactions, GAE~\cite{GAE} leverages the model’s attention layers to produce relevance maps for each of the interactions between the input modalities in the network and it is adaptable to any Transformer-based architectures. IIA~\cite{IIA} iteratively accumulates information from interpolated internal network representations and their gradients to enhance interpretability. Additionally, perturbation-based methods~\cite{RISE,MeaningfulPerburbation} interpret ViT models by applying randomized or meaningful perturbations to input images. These methods~\cite{TransformerLRP,GAE,IIA,RISE,MeaningfulPerburbation} can only generate explanation maps for the final block of ViT models, limiting their ability to reveal the evolution of attention throughout the network. In this work, we address this open problem of interpreting internal attention evolution process inside a ViT by proposing a novel DAAM that extracts and stores the spatial semantic information during the calculation process of the [class] token for explainable visualization. Therefore, the explanation map flow generated by DAAM can correctly highlight the regions on images, on which a ViT model looks at when a decision is progressively formed block by block.

\section{Method}
\label{theProposedMethod}
  \subsection{Preliminaries}
Given a patch size $P$, Vision Transformer~\cite{ViT} firstly splits a 2D image I $\in$ $R^{C \times H \times W}$ into $\frac{H \times W}{P^2}$ patches and then maps each patch into a token $\in R^{D \times 1 \times 1}$ by using a convolution layer, where $H$, $W$, $C$, and $D$ denote the height, width, the number of channels of the input image, and the token dimension respectively. Moreover, all the tokens are flattened in sequence and then concatenated with an extra learnable class token. The visual token sequence $M \in R^{D \times {(1+ \frac{H\times W}{P^2})}}$ is formed by adding  $\frac{H\times W}{P^2}$ learnable position embeddings in an element-wise manner, which can be expressed as,
\begin{equation}
\begin{aligned}
\label{eq1}
M_0=[x_0^0;x_0^1;...;x_0^p;....;x_0^N]+E_{pos},
\end{aligned}
\end{equation}
where $x_0^p \in R^{D \times 1 \times 1}$ is the token of the $p$-th patch if $p>0$, and [class] token is denoted as $x_0^0$. $N$ denotes the number of patches, which is $N=\frac{H\times W}{P^2}$. $E_{pos}$ represents the learnable position embeddings. The token sequence $M$ is sequentially fed into $N$ cascaded transformer blocks ($B$ is the number of transformer blocks), each block consists of a multi-head self-attention (MHSA) module and a Feed Forward Neural Network (FFN). In MHSA module of the $b$-th block ($1\le b \le B$), the token sequence $M_{b-1}$ is fed into a Layer Normalization layer, then the output is projected into a query matrix $Q_b \in R^{(N+1)\times D}$, a key matrix $K_b \in R^{(N+1)\times D}$ and a value matrix $V_b \in R^{(N+1)\times D}$, respectively. Thus, the self-attention matrix $A_b \in R^{(N+1)\times (N+1)}$ is calculated as follows,
\begin{equation}
\begin{aligned}
\label{eq2}
A_b=\text{Softmax}\left( \frac{Q_bK_b^T}{\sqrt{D}} \right)=[a_b^0;a_b^1;...;a_b^N].
\end{aligned}
\end{equation}
Because multi-head mechanism does not impact our proposed method, for simplicity, it is not introduced in this subsection, we can use $A_b$ in Eq.~(\ref{eq2}) to represent self-attention matrix. The $a_b^0 \in R^{1\times(N+1)}$ is called class attention, which reflects the interaction relation between [class] token $x_b^0$ and all patch tokens. The self-attention matrix $A_b$ is multiplied by the value matrix $V_b$, which is expressed by the following equation,
\begin{equation}
\begin{aligned}
\label{eq3}
{A_b}{V_b}=\text{Softmax}\left(\frac{Q_bK_b^T}{\sqrt{D}}\right){V_b}.
\end{aligned}
\end{equation}
Next, the ${A_b}{V_b}$ is projected by a linear layer into $D$-dimensional token sequence before the residual connection~\cite{ResNet}. Thus, the output of MHSA module can be expressed by
\begin{equation}
\begin{aligned}
\label{eq4}
\text{Out}_{\text{MHSA}}^b=\text{Linear}({A_b}{V_b})+M_{b-1}.
\end{aligned}
\end{equation}
$\text{Out}_{\text{MHSA}}^b$ is sent into a layer normalization function before going into FFN that consists of two linear layers and an activation function in between. The output from FFN should be added by $\text{Out}_{\text{MHSA}}^b$ due to skip-connection~\cite{ResNet}. The output token sequence $M_b$ from the $b$-th block can be calculated by
\begin{equation}
\begin{aligned}
\label{eq5}
\text{M}_{b}=\text{FFN}(\text{LN}({\text{Out}_{\text{MHSA}}^b}))+\text{Out}_{\text{MHSA}}^b,
\end{aligned}
\end{equation}
where $\text{LN}$ denotes a layer normalization function. The $M_b$ is viewed as the input token sequence for the $(b+1)$-th transformer block. After a series of MHSA-FFN transformations, the [class] token $x_B^0$ from the final output $M_B$ is sent into the FC layer as the classifier to predict the category of input image $I$. 
\begin{figure}[t]
\centering
\includegraphics[width=1.0\textwidth]{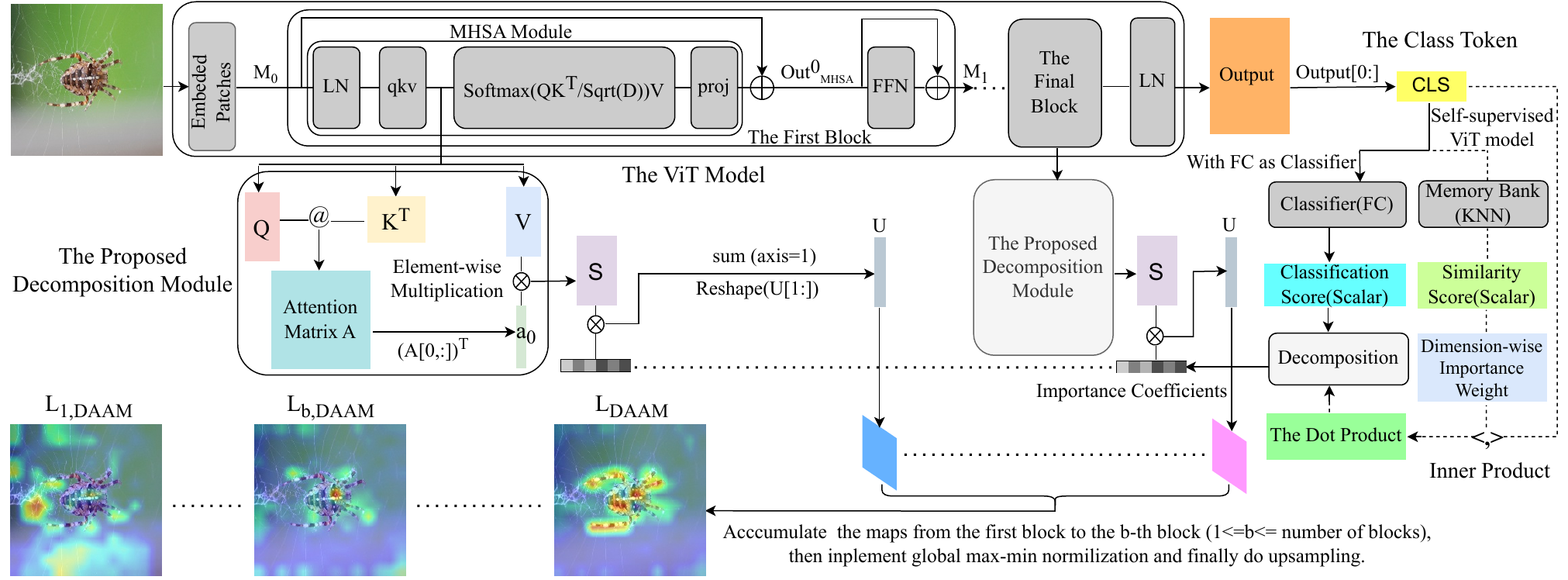}
 \vskip -0.1in
 \caption{The pipeline of the proposed DAAM. The operation @ denotes the matrix multiplication. The dashed line represents an optional pipeline for self-supervised ViT models.~An image is firstly fed into a ViT model to generate the decision-making [class] token. Then, the [class] token is fed into the classifier or memory bank (the dashed line) to calculate the classification score or the inner product between similarity score and the proposed dimension-wise importance weight. The importance coefficients are obtained by decomposing the classification score or the inner product. The semantic feature map extracted by the proposed decomposition module are element-wisely multiplied by the importance coefficients to form the attention map for each intermediate block. The proposed DAAM is generated by accumulating the attention maps from the first block to the final block.}
 \label{fig1}
 \vskip -0.1in
\end{figure}
\subsection{The Proposed Decomposition Module}
As Fig.~\ref{fig1} shows, the proposed module is inserted into each ViT block to extract the spatial information during the calculation of the [class] token. It is worth noting that only [class] token $x_B^0$ plays the role of decision-making~\cite{TransformerLRP}. According to Eqs.~(\ref{eq1})-(\ref{eq5}), the [class] token $x_B^0$ can be derived by 
\begin{equation}\begin{aligned}
\label{eq6}
x_B^0=\text{FFN}(\text{LN}(\text{Linear}(T_B)+x_{B-1}^0))+\text{Linear}(T_B)+x_{B-1}^0,
\end{aligned}\end{equation}
where $T_B=a_B^0V_B$ and $T_B\in R^{1 \times D}$. To interpret why a ViT model predicts the input image to be a specific class, it is crucial to visualize the regions that the [class] token $x_B^0$ utilizes for a specific class. Our proposed method aims to display the spatial feature hidden in the [class] token $x_B^0$. To this end, the spatial information should be exposed firstly from the $B$-th block and then from the $(B-1)$-th class token $x_{B-1}^0$. According to Eq.~(\ref{eq6}), $a_B^0V_B$ is a matrix multiplication between the class attention $a_B^0$ and the value matrix $V_B$ from the $B$-th block. In order to conserve spatial information, we replace it with element-wise product. In other words, $V_B$ can also be seen as $D$ vectors of $(N+1)$-dimension. Because $a_B^0$ is actually a $(N+1)$-dimensional vector, $a_B^0$ is element-wise multiplied by $D$ vectors of $(N+1)$-dimension respectively. The result is a matrix $S_B \in R^{(N+1)\times D }$, which is expressed by 
\begin{equation}\begin{aligned}
\label{eq7}
S_B=a_B^0 \otimes V_B,
\end{aligned}\end{equation}
where $\otimes$ denotes element-wise multiplication, so $T_B=\sum_1^{N+1}{S_B}$. Thus, $S_B$ and $V_B$ are the matrices with the same height and width. Each element of the [class] token $x_B^0$ is the sum of $(N+1)$ elements from $S_B$, or it can be seen as the weighted sum of $(N+1)$ elements of a $(N+1)$-dimensional vector in $V_N$ with the attention weights from the class attention $a_B^0$. The distribution of N+1 vectors of $D$-dimension in $S_B$ can reflect the spatial feature information that the class token utilizes for class prediction. The pipeline for the proposed DAAM is illustrated in Fig.~\ref{fig1}.

For the ViT models with a FC layer as the classifier, classification score is used for making decision. Given a classification score $Y^c$ for a specific class $c$, the classification score can be expressed by $Y^c={X^0_B} {W^c}+\text{bias}$, where $W^c$ denotes the weight vector indicated by the class $c$ from the FC layer, $\text{bias}$ denotes the bias from the FC layer. By decomposing $Y^c$, the channel-wise importance coefficients $C_B^c \in R^{1\times D}$ assigned to each element in $T_B$ can be obtained by
\begin{equation}\begin{aligned}
\label{eq9}
C_B^c=\frac{\partial (Y^c-\text{bias})}{\partial{X_B}^0}\frac{\partial{X_B^0}}{\partial{T_B}}=W^c\frac{\partial{X_B^0}}{\partial{T_B}},
\end{aligned}\end{equation}
where $\frac{\partial{X_B^0}}{\partial{T_B}}$ can be calculated according to Eq~\ref{eq6}. To exclude the negative impact on classification score, $C_B$ is fed into a ReLU~\cite{ReLU} function, and then the output of ReLU function is element-wise multiplied with ${{S_B}}$. The spatial attention is obtained by summing all channel information together. The sum is a $(N+1)$-dimensional vector. By removing the first element from the $(N+1)$-dimensional vector, the attention map $L_B$ for the $B$-th transformer block can be obtained by reshaping the remaining $N$ elements for the size $\frac{H}{P} \times \frac{W}{P}$. The procedure can be described as 
\begin{equation}\begin{aligned}
\label{eq8}
L_B^c=\text{Reshape}\left((\sum_{d=0}^{D-1}{\text{ReLU}({C_B^c})\otimes{S_B}})[1:]\right),
\end{aligned}\end{equation}
where $L_B^c \in R^{\frac{H}{P} \times \frac{W}{P}}$. According to Eq.~(\ref{eq8}), $L_B^c$ reveals the spatial feature information learned by the $B$-th Block. From Eq.~(\ref{eq6}), the spatial feature information hidden in $x_{B-1}^0$ isn't exposed. To this end, by recursively replacing $B$ with $B-1$, $B-2$,..., $1$, the following equation can be obtained,
\begin{equation}\begin{aligned}
\label{eq11}
x_B^0=\sum_{i=1}^B\text{FFN}(\text{LN}(\text{Linear}(T_i) + X^0_{i-1})) +\sum_{i=1}^B\text{Linear}(T_{i}).\\
\end{aligned}\end{equation}
We repeatly perform operations in Eqs.~(\ref{eq6}),~(\ref{eq7}),~(\ref{eq8}),  (\ref{eq9}) and (\ref{eq11}) by replacing $B$ with $B-1$, $B-2$,..., $1$, respectively. Hence, $L_{B-1}^c$,$L_{B-2}^c$,..., and $L_{1}^c$ are obtained respectively. According to Eq.~(\ref{eq11}), the [class] token $x_B^0$ should use the spatial feature information learned by all the blocks to predict the class $c$. Therefore, the attention map for the specific class $c$ should be the sum from $L_1^c$ to $L_B^c$, as 
\begin{equation}\begin{aligned}
\label{eq12}
L^c_{\text{DAAM}}=\sum_{b=1}^{B}L_b^c.
\end{aligned}\end{equation}
To generate the attention map for the intermediate $\bar{B}$-th block ($1\le \bar{B} < B$), based on Eq.~\ref{eq12}, we have
\begin{equation}\begin{aligned}
\label{eq13}
L^c_{\bar{B},\text{DAAM}}=\sum_{b=1}^{\bar{B}}L_b^c.
\end{aligned}\end{equation}

\subsection{The Proposed Dimension-Wise Importance Weight}
 For self-supervised ViT models (e.g., DINO~\cite{DINO} and XCiT~\cite{XCiT}), the memory bank mechanism~\cite{NPID} is usually chosen to make the prediction. We assume that the [class] token $x_B^0$ are similar to $K$ feature vectors in the memory bank. Each feature vector is denoted as $z_k (1 \le k \le K)$. Both $x_B^0$ and $z_k$ have the same number of dimension $D$, so $x_B^0=[\chi_1,\chi_2,...,\chi_d,...,\chi_D]$ and $z_k=[\nu_1^k,\nu_2^k,...,\nu_d^k,...,\nu_D^k]$. As the cosine similarity is used as the KNN metric in memory bank,  the decision-making similarity can be expressed as $\text{cos}(x_B^0,z_k)$. We can compute the dimension-wise importance weight to the similarity by the following equation,
\begin{equation}
\begin{aligned}
\label{eq14}  
w_d=\frac{1}{K}{\sum_{k=1}^{K}\frac{{\chi_d} \times {\nu_d^k}}{ {\vert\vert{\text{cos}(x_B^0,z_k)}\vert\vert}\cdot {\vert\vert{x_B^0}\vert\vert}\cdot{\vert\vert{z_k}\vert\vert}  }},
d=1, 2,\cdots,D,
\end{aligned}
\end{equation}
where $\vert\vert\cdot\vert\vert$ denotes $L^2$-norm. The vector $W$ ($W=[w_1,w_2,...,w_d,...,w_D]$) can be used to replace the weight $W^c$ in Eq.~(\ref{eq9}). Then, combining Eq.~(\ref{eq14}) with Eqs.~(\ref{eq8}) and (\ref{eq12}), our proposed DAAM for self-supervised ViT models can be expressed by 
\begin{equation}
\begin{aligned}
\label{eq16}
L_{\text{DAAM}}=\sum_{b=1}^{B}\text{Reshape}\left( (\sum_{d=1}^{D}\text{ReLU}(w_d) \otimes S_b)[1:]\right).\\
\end{aligned}
\end{equation}

Notably, the bilinear interpolation is adopted as the up-sampling technique for visualisation of $L^c_{\text{DAAM}}$ or $L_{\text{DAAM}}$.
\section{Experiment}
\subsection{Dataset}
Experiments are conducted on the ImageNet dataset (ILSVARC2012)~\cite{ImageNet2012}, which contains around $1.3$ million training images and $50,000$ images in validation set, labelled across $1,000$ semantic categories. The quantitative and qualitative evaluation for the proposed method is performed on the validation set. The validation set provides bounding box annotations for each image, which are useful when evaluating localization capacity of the proposed method.
\subsection{Models and Implementation Details}
Nine ViT models including seven widely-used ViTs with the FC layers as the classifier (DeiT-small-patch16~\cite{DeiT}, DeiT-tiny-patch16~\cite{DeiT}, CaiT-S24~\cite{CaiT}, T2T-ViT-14~\cite{T2TViT}, T2T-ViT-24~\cite{T2TViT}, ViT-small~\cite{ViT}, and ViT-base~\cite{ViT}) and two famous self-supervised ViTs (DINO-ViT-small-patch8~\cite{DINO} and XCiT-m24~\cite{XCiT}) are selected for evaluating the effectiveness of the proposed method. Our implementation is based on Pytorch image models library (Timm). All images in the validation set are resized into $224 \times 224$ pixels first and then all pixels are scaled into [0, 1]. Before feeding into ViT models, they are normalized by using mean [0.485,0.456,0.406] and standard deviation [0.229, 0.224, 0.225]. No further data augmentation is used in our experiments. The quantitative analysis is performed on the correct predictions. All experiments are implemented by using Pytorch library with Python 3.8 on NVIDIA RTX 3090 GPU. 


\subsection{The Attention Flow for Various ViT Models}
\begin{figure*}[h!]
\centering
\includegraphics[width=1.0\textwidth]{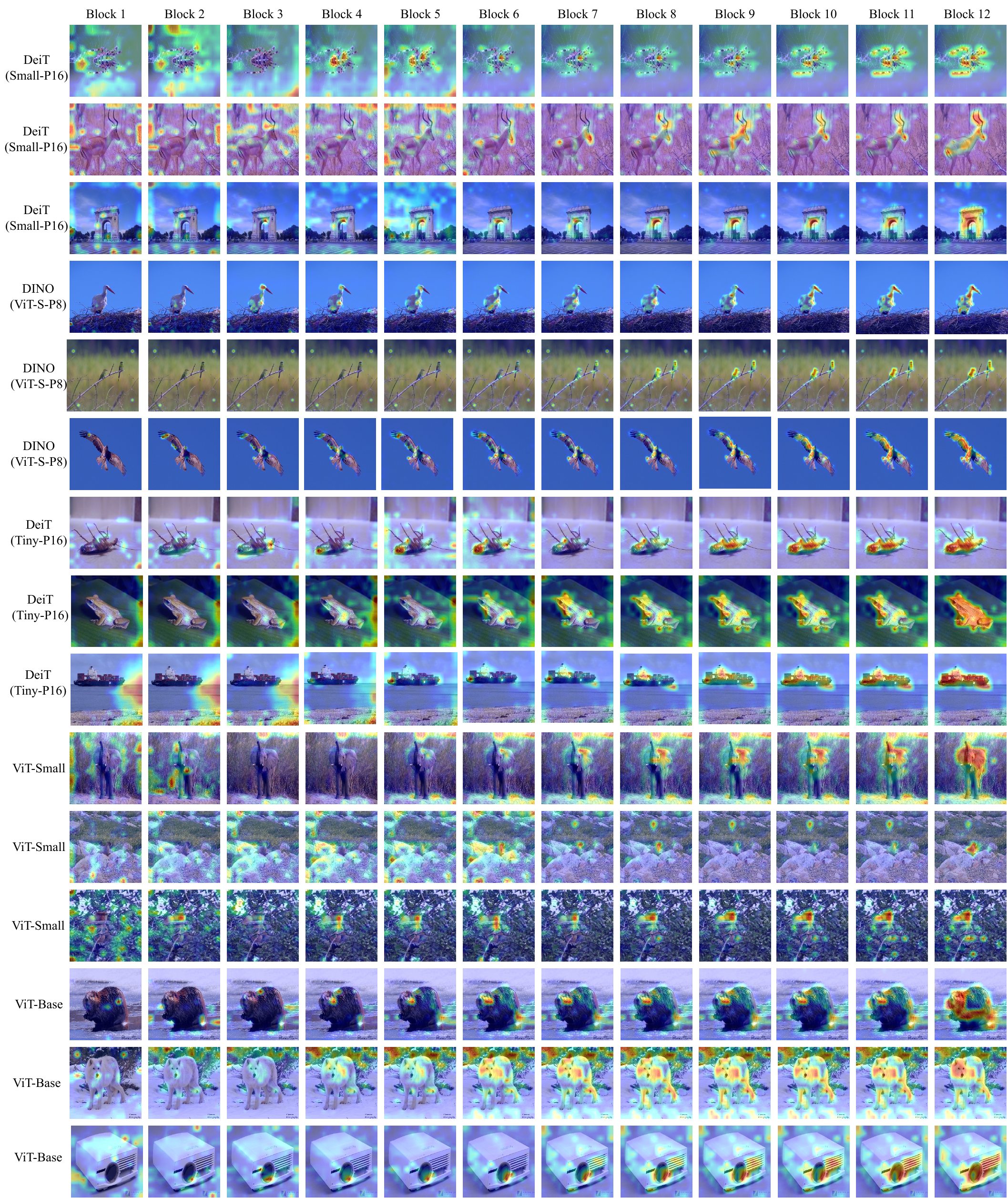}
\caption{The attention flows generated by the proposed DAAM for five pretrained ViT models, DeiT-small-patch16~\cite{DeiT}, DeiT-tiny-patch16~\cite{DeiT}, DINO-ViT-small-patch8~\cite{DINO}, ViT-small~\cite{ViT}, and ViT-base~\cite{ViT}.}
\vspace{-0.5cm}
\label{fig4}
\end{figure*}

\begin{figure*}[t]
\centering
\includegraphics[width=1.0\textwidth]{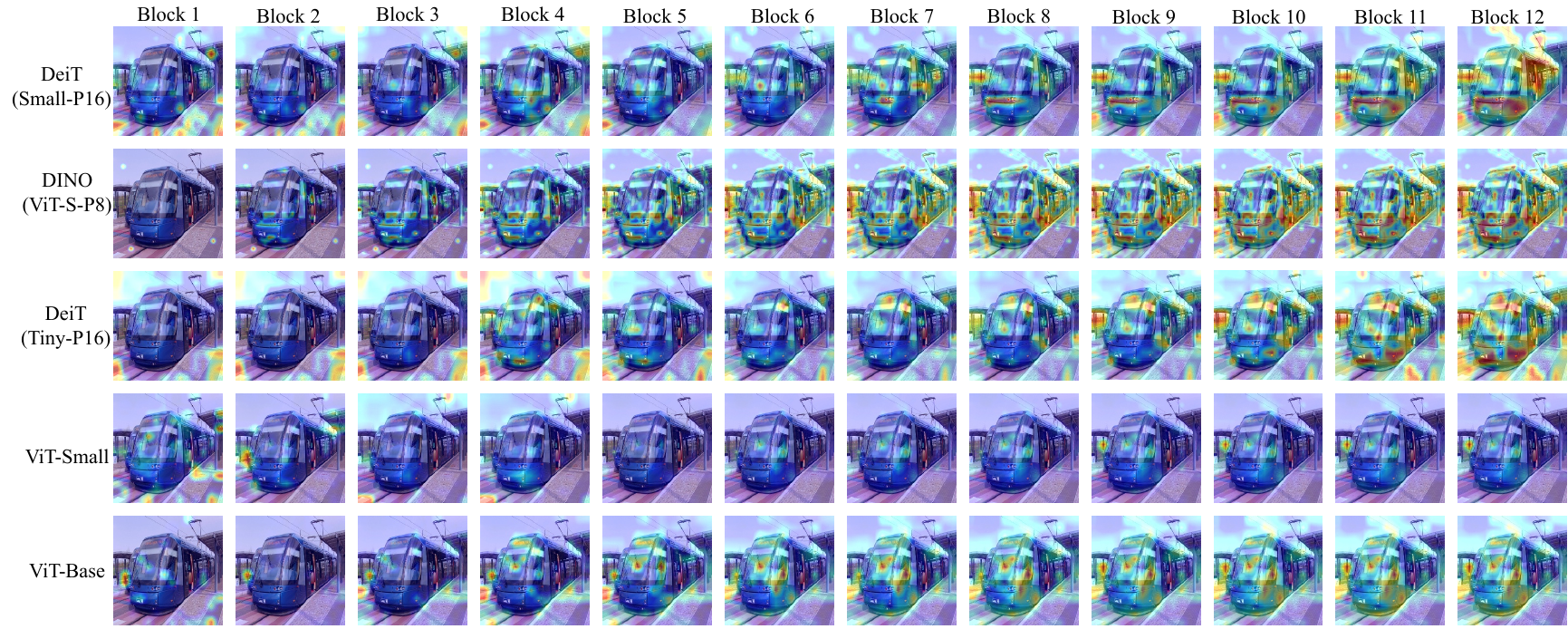}
 \vskip -0.1in
 \caption{{The attention flow generated by the proposed DAAM for DeiT-small-patch16~\cite{DeiT}, DINO-ViT-small-patch8~\cite{DINO}, DeiT-tiny-patch16~\cite{DeiT}, ViT-small~\cite{ViT}, and ViT-base~\cite{ViT} on the same image.}}
 \label{fig_oneImage}
 \vskip -0.1in
\end{figure*}

\begin{sidewaysfigure}
\centering
\includegraphics[width=1.\textwidth]{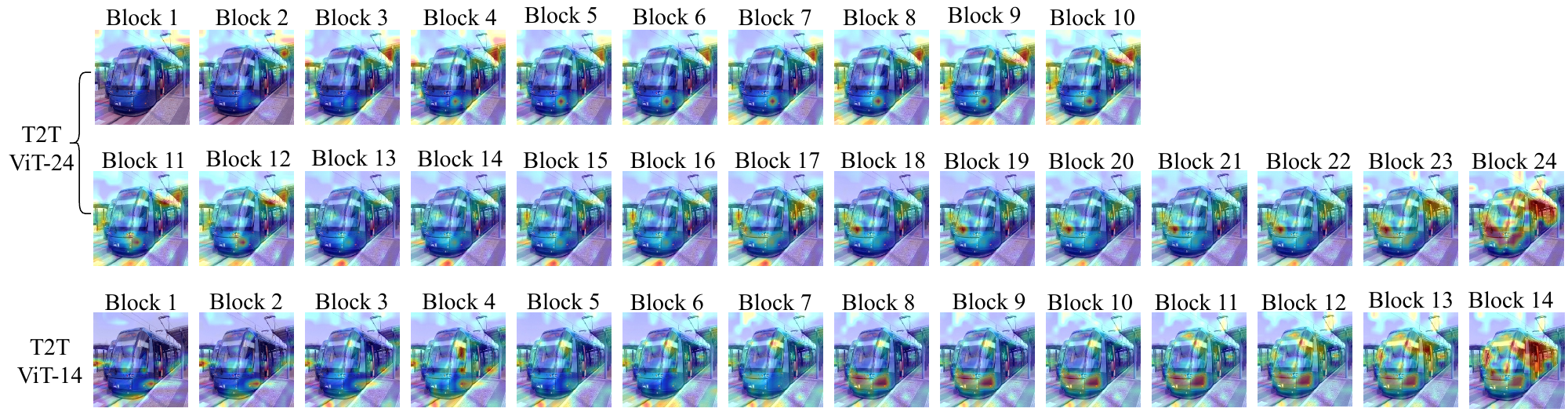}
\caption{The attention flows generated by the proposed DAAM for ViT models of T2T-ViT-14~\cite{T2TViT} and T2T-ViT-24~\cite{T2TViT}.}
\label{figT2TViT}
\end{sidewaysfigure}

\begin{figure*}[t]
\centering
\includegraphics[width=1.0\textwidth]{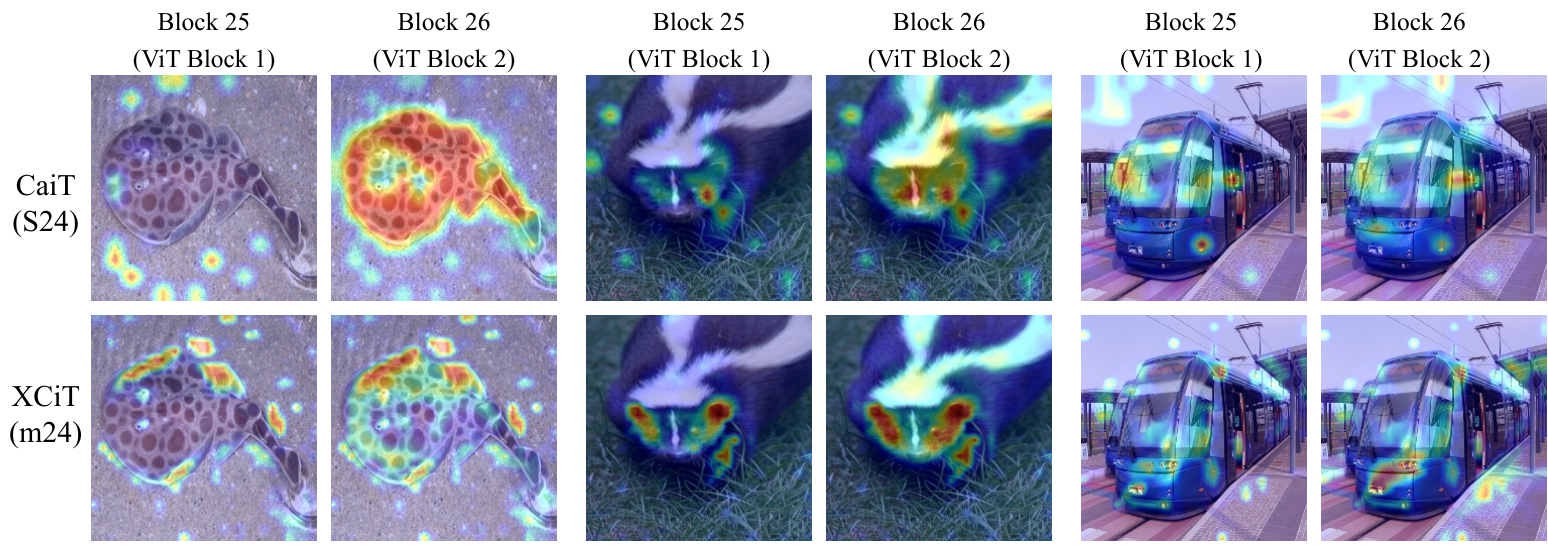}

\caption{{The attention flows generated by the proposed DAAM for two pretrained ViT models of CaiT-S24~\cite{CaiT} ($2$ blocks) and XCiT-m24~\cite{XCiT} ($2$ blocks).}}

\label{fig7}
\end{figure*}

The proposed DAAM is applied to visualise the attention flows from (inside) intermediate blocks to (outside) output block in nine ViT models, including the supervised ViT models (DeiT-small-patch16~\cite{DeiT}, DeiT-tiny-patch16~\cite{DeiT}, CaiT-S24~\cite{CaiT}, T2T-ViT-14~\cite{T2TViT}, T2T-ViT-24~\cite{T2TViT}, ViT-small~\cite{ViT} and ViT-base~\cite{ViT}) and the self-supervised ViT models (DINO-ViT-small-p8~\cite{DINO} and XCiT-m24-p8~\cite{XCiT}). The released pretrained models of DeiT-small-patch16\footnote{https://github.com/facebookresearch/deit}, DeiT-tiny-patch16\footnotemark[1], CaiT-S24-224\footnotemark[1]{}, T2T-ViT-t-14\footnote{https://github.com/yitu-opensource/T2T-ViT}, T2T-ViT-t-24\footnotemark[2]{}, ViT-base\footnote{https://github.com/iia-iccv23/iia}, ViT-small\footnotemark[3]{},DINO-ViT-small-patch8\footnote{https://github.com/facebookresearch/dino}, and XCiT-m24\footnote{https://github.com/facebookresearch/xcit} are used to conduct experiments on the ImageNet2012 validation set, with their reported accuracies of $78.98\%$, $72.20\%$, $83.50\%$,  $81.70\%$, $82.60\%$, $77.91\%$, $83.73\%$, $78.30\%$(KNN), and $77.90\%$(KNN) respectively. Therefore, $39,488$ ($50,000 \times 0.7898$), $36,100$ ($50,000\times 0.7220$), $41,750$ ($50,000 \times 0.8350$), $40,850$ ($50,000 \times 0.8170$), $41,300$ ($50000\times 0.8260$), $38,955$ ($50,000\times0.7791$), $41,865$($50,000\times0.8373$), $39,150$ ($50,000 \times 0.783$), and $38,950$ ($50,000 \times 77.9\%$) testing images from the validation set are correctly predicted by their respective ViT models. Figs.~\ref{fig4},~\ref{fig_oneImage},~\ref{figT2TViT},~\ref{fig7} show the DAAM attention flows of example images that are correctly classified by the ViT models.

As shown in the first row in Fig.~\ref{fig4}, the attention first does not appear on the spider at the first block. Then, the attention gradually transfers to the back of the spider at the block $6$. From the block $7$ onwards, the attention on the spider’s back increases more rapidly and the area of the attention is enlarged to the two hind legs from the block $10$. Using the proposed DAAM, we not only know the features used by a ViT for its decision-making at the final output block, that is the block $12$ in Fig.~\ref{fig4}, but can also find out the contribution of each block to the features in making its prediction.

The proposed DAAM can be used to discern how different ViT models process the same image by comparing the visualizations of the attention maps across all models, as shown in Fig.~\ref{fig_oneImage}, Fig.~\ref{figT2TViT}, and Fig.~\ref{fig7}. Specifically, $5$ ViT models (DeiT-small-patch16, DINO-ViT-small-patch8, DeiT-tiny-patch16, ViT-small, and ViT-base) all include $12$ ViT blocks. $12$ attention maps from the first block to the final block of the $5$ ViT models are visualized from the left to right in Fig.~\ref{fig_oneImage}. It is observed that the $5$ ViT models work differently, as demonstrated by different internal DAAMs. T2T-ViT-14 and T2T-ViT-24 include $14$ and $24$ ViT blocks respectively. The proposed DAAM method can be used to analyze the impact of different number of ViT blocks on attention evolution, as shown in Fig.~\ref{figT2TViT}. 

More importantly, the proposed DAAM can be viewed as a qualitative diagnosis tool that is used to analyse the effect of the module inserted into ViT models in terms of forming the attention on objects. XCiT-m24-p8 has $24$ XCA blocks before $2$ ViT blocks and CaiT-S24 includes $24$ LayerScale blocks, followed by $2$ ViT blocks. It is worth noting that the [class] token is inserted before $2$ ViT blocks rather than before $24$ XCA module or $24$ LayerScale blocks. Hence, attention flow of only two ViT blocks attention flow are visualized for each testing image, as Fig.~\ref{fig7} shows. According to the attention flows shown in Fig.~\ref{fig7}, it can be observed that $24$ XCA modules can help the ViT block to focus on the target object more rapidly than $24$ LayerScale modules on the same image. The proposed DAAM can visually reveal the attention impact caused by the XCA modules and LayerScale modules. Therefore, the proposed DAAM can be used to analyse the effect of any internal component for designing or improving a ViT model. Meanwhile, according to Fig.~\ref{fig7}, when different ViT models make the correct prediction on the same image, they would use different features on the target object to make decision. Different features should reflect different attention evolution pattern because of the diverse architecture designs. Therefore, the attention flow should vary with the architectures of ViT models for the same image. The proposed method is capable of generating explanation maps for intermediate ViT blocks with the availability of the [class] token. However, it currently cannot interpret and visualize the attention flow within intermediate blocks of a ViT where the [class] token is not available. To address this limitation, future research will focus on leveraging the entire attention matrix to preserve spatial information. This advancement is expected to broaden the applicability of the proposed method across a wider range of ViT architectures.
\begin{figure}[t]
\centering
\includegraphics[width=1.0\columnwidth]{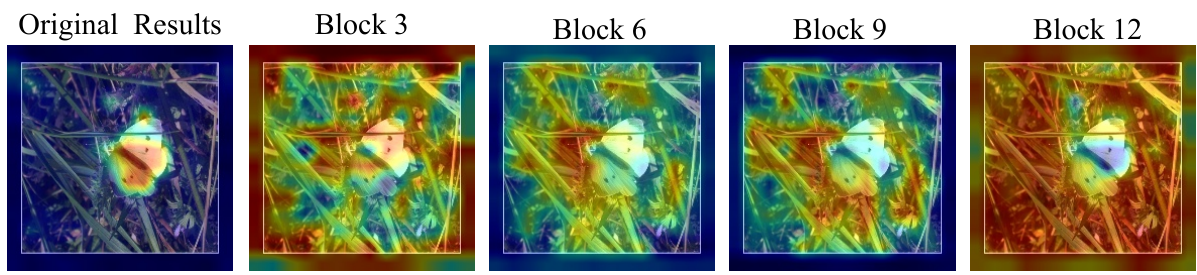}
\caption{Sanity check for the proposed DAAM by cascading randomization from top block to block $3$, block $6$, block $9$, and block $12$ in DeiT-small-patch16~\cite{DeiT} respectively. The original results are generated for the pretrained DeiT-small-patch16 without any parameters randomization.}
\label{fig6}
\end{figure}
\subsection{Sanity Check}
Sanity check aims to evaluate the sensitivity of the attention maps to the parameters of the deep learning models~\cite{SanityCheck}. We evaluate the proposed DAAM by cascading randomization of the  weights of linear layers from MHSA modules in ViT blocks. Fig.~\ref{fig6} shows the sanity analysis for the attention maps from the last block of DeiT-small-patch16~\cite{DeiT} model obtained by progressively randomizing the parameters from the top block to block $3$, block $6$, block $9$, and block $12$ respectively. The results shows that the attention maps are destroyed with the parameters randomization. Thus, the proposed DAAM is sensitive to the ViT model parameters.
\begin{figure}[h!]
\centering
\includegraphics[width=0.72\textwidth]{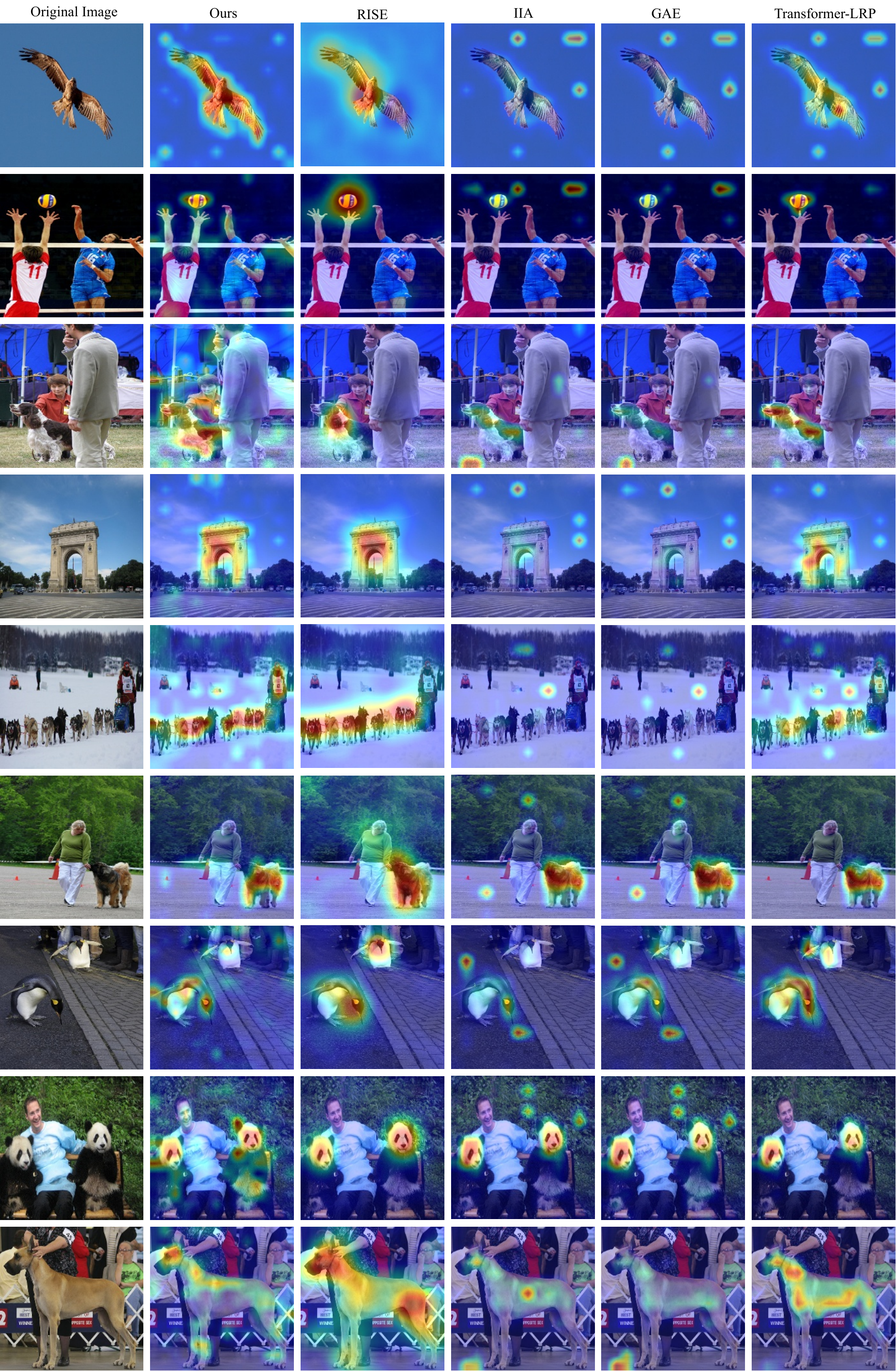}
\caption{{Visual examples of localization capability of the proposed DAAM in comparison with existing methods for the last block in the pretrained model DeiT-small-patch16~\cite{DeiT}. The ground truth label of the images from the top row to the bottom row are Kite (hawk), volleyball, English Springer, Triumphal Arch, dog sled, Leonberg dog, King Penguin, Giant Panda, and Great Dane, respectively.}}
\vspace{-0.5cm}
\label{fig_comparison}
\end{figure}
\subsection{{Qualitative Comparison on the Last Block}}
To evaluate the effectiveness of the proposed DAAM method, a qualitative comparison is performed between the proposed DAAM and RISE~\cite{RISE}, IIA~\cite{IIA}, GAE~\cite{GAE}, and Transformer-LRP~\cite{TransformerLRP}. It must be noted that the proposed DAAM is the first tool that can visually explain all the intermediate blocks of a ViT, while all the existing ViT explanation methods can only explain the final output of the last block. Here we only can compare the attention maps for the last block because no existing explanation methods can explain internal blocks as the proposed DAAM does. For a fair comparison, the visualization experiments are conducted on the same images for the same ViT model of pretrained DeiT-small-patch16. We use the public released code\footnote{{https://github.com/eclique/RISE}} to perform RISE, the released code\footnote{{https://github.com/iia-iccv23/iia}} to perform IIA, the released code\footnote{{https://github.com/hila-chefer/Transformer-MM-Explainability}} to perform GAE, and the released code\footnote{{https://github.com/hila-chefer/Transformer-Explainability}} to perform Transformer-LRP, respectively. Fig.~\ref{fig_comparison} displays the explanation maps for the last block of the ViT generated by the four explanation methods on the randomly selected images that are correctly classified by the ViT model.

\begin{table}[t]
\caption{The localization capability of the proposed DAAM in comparison with Transformer-LRP~\cite{TransformerLRP}, GAE~\cite{GAE}, IIA2~\cite{IIA}, and IIA3~\cite{IIA} for ViT-small~\cite{ViT}. For the metrics of AUC Pos Perturb, AUC Del, and ADP on both predicted and target labels, lower is better. For the metrics of AUC Neg Perturb, AUC Ins, PIC on both predicted and target labels, higher is better. The best result is highlighted by using the bold font.}
\label{table3}
\centering
\scalebox{0.80}{
\begin{tabular}{llccccc}
\toprule
\multicolumn{2}{c}{ViT-small}&T-LRP~\cite{TransformerLRP} &GAE~\cite{GAE}& IIA2~\cite{IIA}&IIA3~\cite{IIA}&Ours\\
\midrule
AUC Neg Perturb & Predicted   & $53.29\%$ & $52.81\%$& $55.76\%$& $56.39\%$& $\boldsymbol{57.36\%}$\\
AUC Neg Perturb & Target       & $53.93\%$ & $53.58\%$& $58.71\%$& $\boldsymbol{59.31\%}$& $58.67\%$ \\
AUC Pos Perturb & Predicted    & $14.16\%$ & $14.75\%$& $13.06\%$& $12.15\%$& $\boldsymbol{12.04\%}$\\
AUC Pos Perturb & Target       & $13.08\%$ & $14.38\%$& $12.97\%$& $\boldsymbol{11.86\%}$& $11.87\%$\\
AUC Ins         & Predicted    & $45.42\%$ & $45.21\%$& $46.55\%$& $47.68\%$& $\boldsymbol{48.44\%}$\\
AUC Ins         & Target       & $46.12\%$ & $45.69\%$& $47.83\%$& $48.53\%$& $\boldsymbol{48.95\%}$\\
AUC Del         & Predicted    & $11.28\%$ & $11.92\%$& $11.18\%$& $\boldsymbol{10.31\%}$& $10.34\%$\\
AUC Del         & Target       & $11.06\%$ & $11.69\%$& $10.98\%$& $10.16\%$& $\boldsymbol{9.98\%}$\\
PIC             & Predicted       & $13.67\%$ & $8.68\%$&  $15.49\%$& $17.79\%$& $\boldsymbol{22.16\%}$\\
PIC             & Target       & $15.00\%$ & $10.02\%$& $18.14\%$& $19.59\%$& $\boldsymbol{24.23\%}$\\
ADP             & Predicted    & $51.94\%$ & $36.98\%$& $36.74\%$& $36.04\%$& $\boldsymbol{32.29\%}$\\
ADP             & Target       & $50.59\%$ & $64.72\%$& $39.58\%$& $39.43\%$& $\boldsymbol{35.99\%}$\\
\bottomrule
\end{tabular}}
\end{table}
\begin{table}[h!]
\caption{The localization capability of the proposed DAAM in comparison with Transformer-LRP~\cite{TransformerLRP}, GAE~\cite{GAE}, IIA2~\cite{IIA}, and IIA3~\cite{IIA} for ViT-base~\cite{ViT}. For the metrics of AUC Pos Perturb, AUC Del, and ADP on both predicted and target labels, lower is better. For the metrics of AUC Neg Perturb, AUC Ins, PIC on both predicted and target labels, higher is better. The best result is highlighted by using the bold font.}
\label{table2}
\centering
\scalebox{0.80}{
\begin{tabular}{llccccc}
\toprule
\multicolumn{2}{c}{ViT-base}  & T-LRP~\cite{TransformerLRP} &GAE~\cite{GAE}& IIA2~\cite{IIA}& IIA3~\cite{IIA}&Ours\\
\midrule
AUC Neg Perturb & Predicted   & $54.16\%$ & $54.61\%$& $56.01\%$& $57.68\%$& $\boldsymbol{57.85\%}$\\
AUC Neg Perturb & Target          & $55.04\%$ & $55.67\%$& $57.47\%$& $58.31\%$& $\boldsymbol{59.10\%}$ \\
    AUC Pos Perturb & Predicted    & $17.03\%$ & $17.32\%$& $15.19\%$& $14.96\%$& $\boldsymbol{14.09\%}$\\
    AUC Pos Perturb & Target          & $16.04\%$ & $16.72\%$& $15.81\%$& $15.02\%$& $\boldsymbol{14.92\%}$\\
    AUC Ins         & Predicted        & $48.58\%$ & $48.96\%$& $49.31\%$& $50.71\%$& $\boldsymbol{51.06\%}$\\
    AUC Ins         & Target          & $49.19\%$ & $49.65\%$& $50.49\%$& $51.26\%$& $\boldsymbol{51.71\%}$\\
    AUC Del         & Predicted        & $14.20\%$ & $14.37\%$& $12.89\%$& $\boldsymbol{12.25\%}$& $12.90\%$\\
    AUC Del         & Target          & $13.77\%$ & $13.99\%$& $13.12\%$& $\boldsymbol{12.38\%}$& $13.21\%$\\
    PIC             & Predicted       & $13.3\%7$ & $23.65\%$& $26.18\%$& $30.41\%$& $\boldsymbol{32.84\%}$\\
    PIC             & Target          & $14.97\%$ & $25.53\%$& $28.97\%$& $31.75\%$& $\boldsymbol{35.23\%}$\\
    ADP             & Predicted      & $54.02\%$ & $37.84\%$& $33.93\%$& $34.05\%$& $\boldsymbol{32.73\%}$\\
    ADP             & Target          & $56.68\%$ & $36.09\%$& $31.08\%$& $32.64\%$& $\boldsymbol{28.96\%}$\\
\bottomrule
\end{tabular}}
\end{table}

\subsection{Quantitative Analysis of Localization Capacity}
Here, we perform a qualitative evaluation on the localization capacity of the proposed DAAM using the attention map of the final output block, in comparison with the current visual explanation methods for transformers: Transformer-LRP~\cite{TransformerLRP}, GAE~\cite{GAE}, and IIA~\cite{IIA}. This comparison is conducted only on the last frame of the DAAM attention flow video that shows the evolution of attention from inside (the input block, intermediate blocks) to outside (output block) through a ViT architecture, because all existing methods can only explain the attention from outside of output block not inside a ViT.  Using the same experiment setup as in~\cite{IIA}, the comparison results for ViT-small and ViT-base are shown in Table~\ref{table3} and Table~\ref{table2}. As the Table~\ref{table3} shows, the proposed method is superior to~\cite{TransformerLRP,GAE,IIA} in AUC Neg Perturb (predicted), AUC Pos Perturb (predicted), ADP, PIC, AUC Ins, and AUC Del (target). For AUC Neg Perturb (target), AUC Pos Perturb (target), and AUC Del (predicted), the proposed method is better than~\cite{TransformerLRP} and~\cite{GAE}, slightly worse than~\cite{IIA}. They demonstrate that the proposed DAAM generates the most accurate explanation maps in terms of class discrimination and localization. 

\subsection{Comparison with CAM-based Methods} 
The CAM-based methods~\cite{GradCAM,GradCAM++,XGrad-CAM,LayerCAM, ScoreCAM,AblationCAM} are developed to generate the explanation maps~\cite{spare} for CNN models that innately keep spatial information, but fail to work on explaining ViTs, which do not provide the spatial information that CAM-based methods require. However, in the research community, CAM-based methods are widely used to visually explain ViT models using the Pytorch library\footnote{https://github.com/jacobgil/pytorch-grad-cam/blob/master/tutorials/vision\_transformers.md}
\begin{table}[t]
\caption{The comparison of localization capacity between the proposed DAAM and $6$ CAM-based methods for DeiT-small-patch16~\cite{DeiT}, CaiT-S24-224~\cite{CaiT}, and T2T-ViT-t-14~\cite{T2TViT}. Higher value is better for average IoU and Accuracy (IoU$\ge 0.5$).}
\label{table4}
\centering
\scalebox{0.90}{
\begin{tabular}{lcccccc}
\toprule
\multirow{2}{*}{Methods}&\multicolumn{2}{c}{DeiT}&\multicolumn{2}{c}{CaiT}&\multicolumn{2}{c}{T2T-ViT}\\
\cmidrule{2-7}
& IoU & Acc & IoU & Acc&IoU&Acc\\
\midrule
Grad-CAM        & $53.72\%$&$56.40\%$&$41.11\%$&$39.84\%$&$55.35\%$&$59.69\%$\\
Grad-CAM++      & $44.24\%$&$41.47\%$&$35.54\%$&$31.29\%$&$52.57\%$&$54.92\%$\\
XGrad-CAM       & $35.31\%$&$26.71\%$&$37.70\%$&$30.73\%$&$39.45\%$&$32.60\%$\\
Layer-CAM       & $18.09\%$&$8.73\%$&$22.88\%$&$15.85\%$&$29.35\%$&$19.15\%$\\
Ablation-CAM    & $52.33\%$&$52.80\%$&$38.82\%$&$35.94\%$&$48.51\%$&$46.54\%$\\
Score-CAM       & $51.85\%$&$53.86\%$&$38.05\%$&$36.01\%$&$52.90\%$&$54.73\%$\\
Ours            &  $\boldsymbol{56.45\%}$&$\boldsymbol{61.58\%}$ &$\boldsymbol{42.36\%}$ &$\boldsymbol{40.53\%}$&$\boldsymbol{56.08\%}$&$\boldsymbol{61.25\%}$\\
\bottomrule
\end{tabular}}
\end{table}
\begin{figure*}[h!]
\centering
\includegraphics[width=0.95\textwidth]{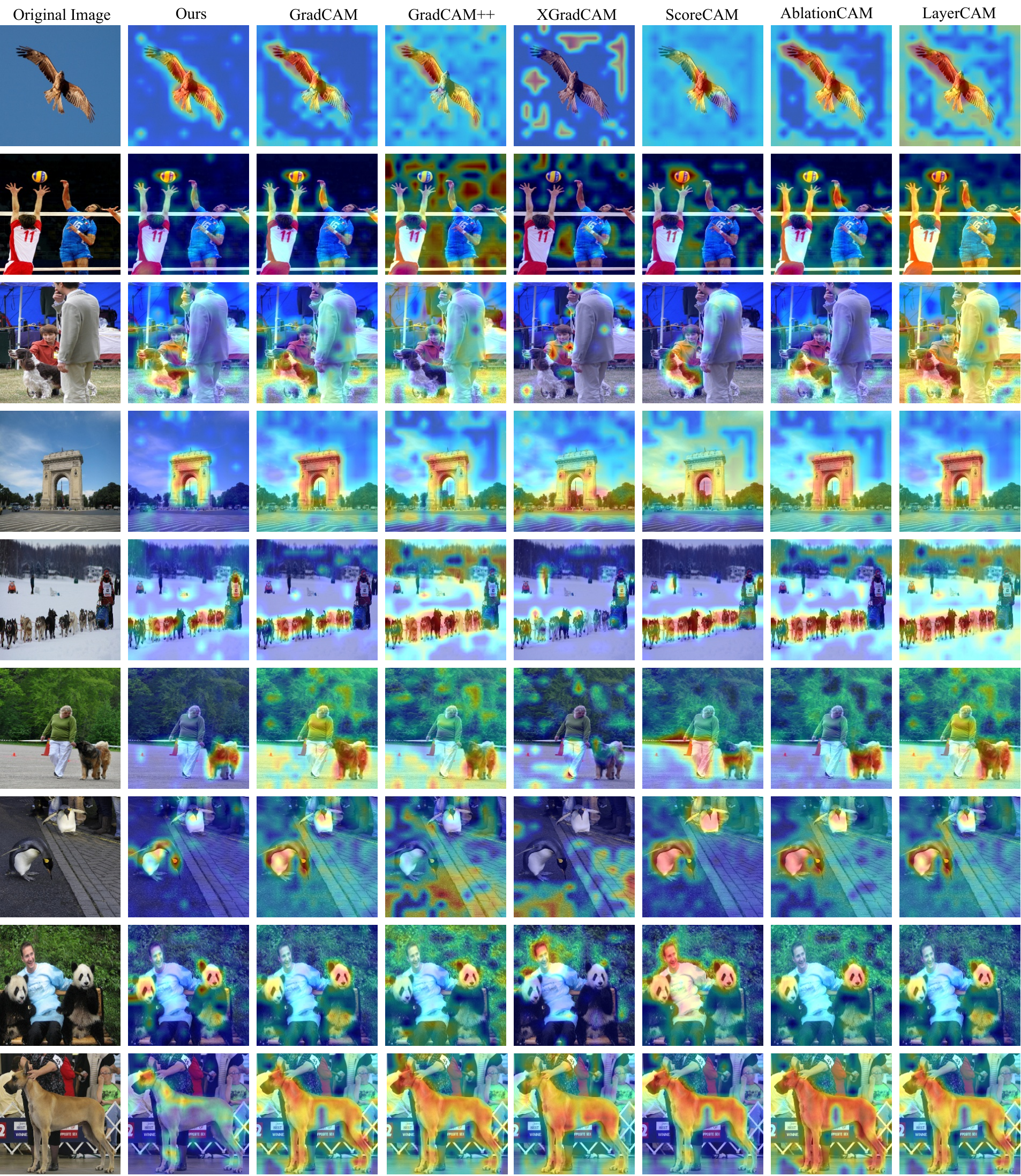}
\caption{The explanation maps generated by the proposed DAAM, GradCAM~\cite{GradCAM}, GradCAM++~\cite{GradCAM++}, XGradCAM~\cite{XGrad-CAM}, ScoreCAM~\cite{ScoreCAM}, AblationCAM~\cite{AblationCAM}, and LayerCAM~\cite{LayerCAM} for the last block in the pretrained model DeiT-small-patch16~\cite{DeiT}. The ground truth label of the image from the top row to the bottom row is Kite (hawk), volleyball, English Springer, Triumphal Arch, dog sled, Leonberg dog, King Penguin, Giant Panda, and Great Dane, respectively.}
\label{figComparewithCAM}
\vskip -0.10in
\end{figure*}
To evaluate the correctness of the proposed DAAM for interpreting ViT models, the quantitative experiments are conducted on DeiT-small-patch16, CaiT-S24-224, and T2T-ViT-t-14 to compare the proposed DAAM with six CAM-based methods. 
The six CAM-based methods include Grad-CAM, Grad-CAM++, XGrad-CAM, Layer-CAM, Ablation-CAM, and Score-CAM. We use the publicly released code\footnote{https://github.com/jacobgil/pytorch-grad-cam} to implement the CAM-based methods. $L_{\text{DAAM}}^c$ defined in Eq.(\ref{eq12}) is used for comparison. The model DeiT-small-patch16, CaiT-S24-224, and T2T-ViT-t-14, which are pretrained well on the ImageNet2012 training set, are adopted to test the ImageNet2012 validation set. The pretrained model DeiT-small-patch16, CaiT-S24-224, and T2T-ViT-t-14 achieve the accuracy of $78.98\%$, $83.50\%$, and $81.70\%$ respectively. Therefore, 39,488 ($50,000 \times 0.7898$), 41,750 ($50,000 \times 0.8350$), and 40,850 ($50,000 \times 0.8170$) testing images from validation set are correctly predicted respectively. The metric intersection over union (IoU)~\cite{LayerCAM} defined in Eq.~(\ref{eq17})  and localization accuracy (IoU is bigger than 0.5)~\cite{LayerCAM} are adopted to measure the localization capacity. The annotated bounding boxes provided by ImageNet2012 validation set are used to compute both metrics. When computing both metrics, all explanation maps are resized to the same size as the original images. IoU definition is
\begin{equation}\begin{aligned}
\label{eq17}
\text{IoU}=\frac{\sum_{(i,j)\in {(\text{bbox}_e \bigcap \text{bbox}_g})}{1}}{\sum_{(i,j)\in{(\text{bbox}_e \bigcup \text{bbox}_g})}{1}},
\end{aligned}\end{equation}
where $\text{bbox}_e$ denotes the estimated bounding box while $\text{bbox}_g$ denotes the ground truth bounding box. To generate an estimated bounding box, the saliency map is binarized with the threshold of the maximum intensity, and then the tight box of the largest connected regions is found as the estimated bounding box~\cite{LayerCAM}. The threshold is set to $0.1$ in our experiments. If IoU is bigger than 0.5 or equal to $0.5$ for a saliency map, the localization of the saliency map is considered as correct, otherwise, the localization is incorrect. Average IoU is calculated over all the correctly predicted samples. The average IoU and accuracy measure different localization capacities. The localization accuracy can reflect the proportion of the saliency maps with correct localization. Table 3 shows the results of average IoU and the accuracy with IoU bigger than $0.5$ for DeiT-small-patch16 (over $39,488$ samples),  CaiT-S24-224 (over $41,750$ samples), and T2T-ViT-t-14 (over $40,850$ samples). From Table~\ref{table4}, the proposed DAAM achieves the best performance via both metrics. 
\begin{figure*}[t]
\centering
\includegraphics[width=1.0\linewidth]{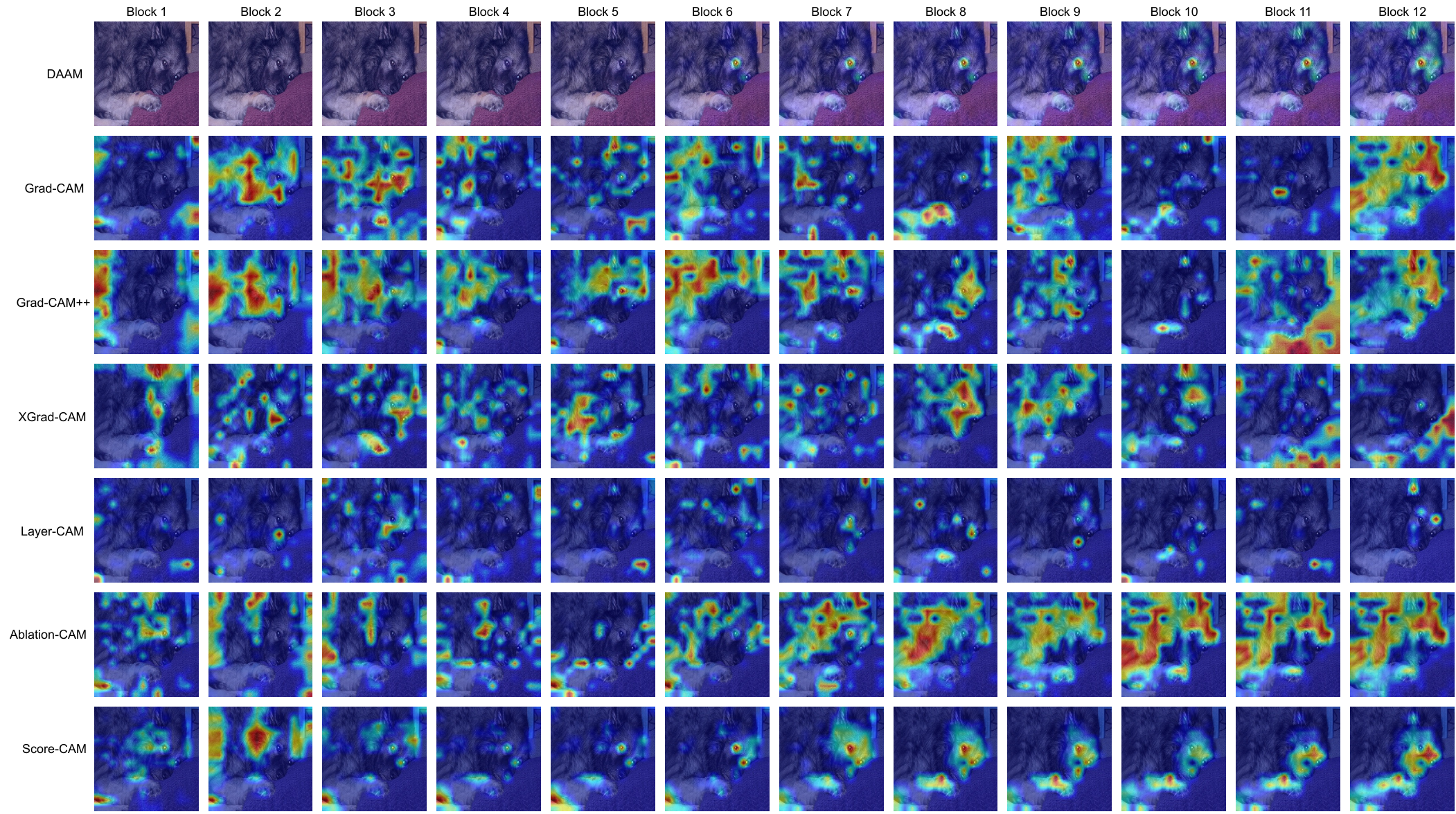}
\caption{Visual comparison of the explanation maps generated by the proposed DAAM and $6$ CAM-based methods for Deit-small-patch16~\cite{DeiT} ($12$ blocks).}
\label{figCompare12maps}
\end{figure*}
To evaluate the superiority of the proposed DAAM, We also conduct the qualitative experiments on DeiT-small-path16 to compare the proposed DAAM with six CAM-based methods. The explanation maps for the last ViT block are shown in Fig.~\ref{figComparewithCAM}. To evaluate the effectiveness of attention evolution revealed by the proposed DAAM, the explanation maps from $12$ ViT blocks of Deit-small-patch16  are generated by the proposed DAAM, which are displayed from left to right on the first row in Fig.~\ref{figCompare12maps}. For fair comparison, $12$ blocks of explanation maps on the same image for Deit-small-patch16 are generated by $6$ CAM-based methods, which are listed in sequence from the second row to the seventh row in Fig.~\ref{figCompare12maps}. The visualization result clearly shows that the proposed DAAM effectively visualizes the evolution process of the attention used for decison-making. No evolution pattern can be observed from the attention maps generated by $6$ CAM-based methods. 
\begin{figure}[t]
\centering
\includegraphics[width=1.0\linewidth]{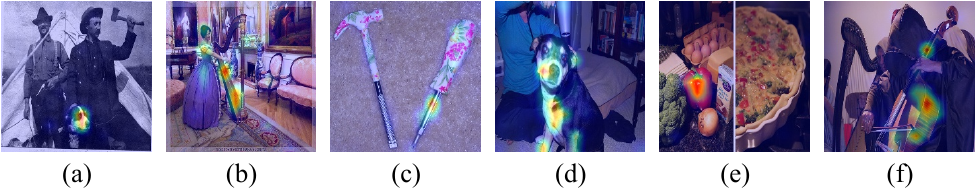}.
\caption{The examples of how DAAM effectively explain incorrect classifications made by the pretrained ViT model (DeiT-small-patch16~\cite{DeiT}). Because the attention of (a), (b), (c), (d), (e), and (f) focuses on the Greater Swiss Mountain Dog, the harp, the screwdriver, the Swiss Mountain Dog, the bell pepper, and the cello respectively, rather than the ground truth labels ((a) hatchet, (b) the hoopskirt, (c) the hammer, (d) the hairdryer, (e) the broccoli, and (f) the harp), the 6 samples from (a) to (f) are classified into the incorrect categories.}
\label{figWrongExamples}
\end{figure}
\subsection{\textit{The Analysis for the Incorrect Classification}}
The proposed DAAM can be used as a visual means to analyze why a ViT model makes incorrect predictions. In our experiment, without retraining on ground truth labels, we applied the proposed DAAM to interpret a ViT, pretrained DeiT-small-patch16 model~\cite{DeiT}, to generate attention maps for six test images that the ViT model misclassified. The results are displayed in Fig.~\ref{figWrongExamples}, which shows
the attention regions generated by the proposed DAAM are consistent with the incorrect prediction made by the DeiT model. For example, Fig.~\ref{figWrongExamples} (a) is classified by the ViT model as the “Greater Swiss Mountain Dog” because the attention visually revealed by DAAM is indeed located on the dog, rather than the “hatchet” (ground truth label of image (a)).

\begin{figure}[t]
\centering
\includegraphics[width=1.0\linewidth]{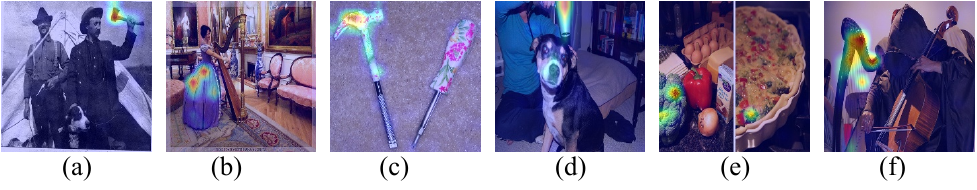}
\caption{The examples of how DAAM effectively locate the target object indicted by the given label for the pretrained DeiT-small-patch16~\cite{DeiT}. Given the ground truth labels from (a) to (f),  the attention generated by DAAM is located on the hatchet on (a), the hoop skirt on (b), the hammer on (c), the hairdryer on (d), the broccoli on (e), and the harp on (f).}
\label{figRecorrect}
\end{figure}

Next, we manually replace the incorrect
prediction label generated by the DeiT model with the ground truth label provided by the ImageNet2012, to see whether the proposed DAAM can discover the
target object consistent with the ground truth label. Let the score vector $[Y^{0},Y^{1},\cdots,Y^{999}]$ denote the output of the pre-trained DeiT-small-patch16 (Note that the ImageNet2012 has 1000 classes). Assuming that the ViT model incorrectly classifies a testing image as class $c_3$, while actually the ground truth label of the testing image is the class $c_5$. In order to visualize what region on the test image is looked at by the ViT model for predicting it as $c_3$, the classification score $Y^{c_3}$ was used by DAAM to generate the explanation map, as shown in Fig. 10. In our experiment, to know what region should be looked at by the ViT model for correctly predicting the test image as $c_5$, $Y^{c_5}$ is used by DAAM to generate the explanation map, as shown in Fig.~\ref{figRecorrect}.  From Fig.~\ref{figRecorrect}(a), it can be seen that if DeiT makes the correct decision of “hatchet”, the proposed DAAM shows the attention the model moves to “hatchet” (see Fig.~\ref{figRecorrect} (a)) from “Greater Swiss Mountain Dog” (see Fig.~\ref{figWrongExamples} (a)). 
The result demonstrates the effectiveness of the proposed DAAM as an interpretive tool for analyzing the incorrect prediction.

\section{Conclusion}
In this work, we propose a novel visual explanation method, Dynamic Accumulated Attention Map (DAAM), to uncover the attention flow of intermediate blocks within ViT models. By leveraging the [class] token that plays a critical role in decision-making, DAAM extracts and preserves spatial semantic features using the self-attention matrix and the value matrix. The importance coefficients are computed by decomposing the classification score. For self-supervised ViT models, these coefficients are derived from the proposed dimension-wise importance weights. The dynamic attention flow is revealed through block-wise accumulation, combining the constructed spatial features with the importance coefficients. 

Extensive quantitative and qualitative experiments demonstrate DAAM's superior localization capability and effectiveness as an interpretative tool. However, the method currently relies on the [class] token. To address this limitation, leveraging the entire attention matrix
to preserve spatial information will be a key focus of future research, thereby extending DAAM's applicability to a wider range of architectures.

Beyond its current applications, DAAM holds significant potential for guiding the design of new classification models, advancing weakly supervised techniques, and enhancing related research fields such as model optimization and adversarial attack defense. Additionally, integrating DAAM with iterative integration-based explanation methods could enable its adaptation to CNN-based models, transforming it into a model-agnostic interpretability tool. These advancements will not only address the current limitations but also broaden the scope and impact of DAAM in both research and practical applications.
\section{Acknowledgement}
This work was supported in part by the Australian Research Council under Discovery Grant DP180100958 and Industrial Transformation Research Hub Grant IH180100002.
\bibliography{DAAM}

\begin{thebibliography}{10}
\expandafter\ifx\csname url\endcsname\relax
  \def\url#1{\texttt{#1}}\fi
\expandafter\ifx\csname urlprefix\endcsname\relax\def\urlprefix{URL }\fi
\expandafter\ifx\csname href\endcsname\relax
  \def\href#1#2{#2} \def\path#1{#1}\fi

\bibitem{RangViT}
A.~Ando, S.~Gidaris, A.~Bursuc, G.~Puy, A.~Boulch, R.~Marlet, Rangevit: Towards vision transformers for 3d semantic segmentation in autonomous driving, in: Proceedings of the IEEE Conference on Computer Vision and Pattern Recognition (CVPR), 2023, pp. 5240--5250.

\bibitem{uavformer}
S.~Yi, X.~Liu, J.~Li, L.~Chen, Uavformer: A composite transformer network for urban scene segmentation of uav images, Pattern Recognition 133 (2023) 109019.

\bibitem{medvit}
O.~N. Manzari, H.~Ahmadabadi, H.~Kashiani, S.~B. Shokouhi, A.~Ayatollahi, Medvit: A robust vision transformer for generalized medical image classification, Computers in Biology and Medicine 157 (2023) 106791.

\bibitem{SMGUNet}
J.~Yan, K.~Zhang, Q.~Sun, C.~Ge, W.~Wan, J.~Sun, H.~Zhang, Spatial-spectral unfolding network with mutual guidance for multispectral and hyperspectral image fusion, Pattern Recognition 161 (2025) 111277.

\bibitem{Transformer}
A.~Vaswani, N.~Shazeer, N.~Parmar, J.~Uszkoreit, L.~Jones, A.~N. Gomez, L.~U. Kaiser, I.~Polosukhin, {Attention is All you Need}, in: Proceedings of the Advances in Neural Information Processing Systems (NeurIPS), Vol.~30, 2017.

\bibitem{skincander}
G.~H. Dagnaw, M.~El~Mouhtadi, M.~Mustapha, Skin cancer classification using vision transformers and explainable artificial intelligence, Journal of Medical Artificial Intelligence 7.

\bibitem{wsssMRI}
H.~Wang, J.~Wu, C.~Wang, M.~Zhao, D.~Zhang, R.~Xu, Ac-cam: Affinity-aware contrast cam for weakly-supervised semantic segmentation on mri brain tumor, in: IEEE International Conference on Bioinformatics and Biomedicine (BIBM), 2024, pp. 3768--3771.

\bibitem{ViT}
A.~Dosovitskiy, L.~Beyer, A.~Kolesnikov, D.~Weissenborn, X.~Zhai, T.~Unterthiner, M.~Dehghani, M.~Minderer, G.~Heigold, S.~Gelly, et~al., An image is worth 16x16 words transformers for image recognition at scale, in: Proceedings of the International Conference on Learning Representations (ICLR), 2020.

\bibitem{DeiT}
H.~Touvron, M.~Cord, M.~Douze, F.~Massa, A.~Sablayrolles, H.~Jegou, Training data-efficient image transformers and distillation through attention, in: Proceedings of the International Conference on Machine Learning (ICML), Vol. 139, 2021, pp. 10347--10357.

\bibitem{T2TViT}
L.~Yuan, Y.~Chen, T.~Wang, W.~Yu, Y.~Shi, Z.-H. Jiang, F.~E. Tay, J.~Feng, S.~Yan, Tokens-to-token vit: Training vision transformers from scratch on imagenet, in: Proceedings of the IEEE International Conference on Computer Vision (ICCV), 2021, pp. 558--567.

\bibitem{CrossViT}
C.~Chen, Q.~Fan, R.~Panda, Crossvit: Cross-attention multi-scale vision transformer for image classification, in: Proceedings of the IEEE International Conference on Computer Vision (ICCV), 2021, pp. 357--366.

\bibitem{DynamicViT}
Y.~Rao, W.~Zhao, B.~Liu, J.~Lu, J.~Zhou, C.-J. Hsieh, Dynamicvit: Efficient vision transformers with dynamic token sparsification, in: Proceedings of the Advances in Neural Information Processing Systems (NeurIPS), Vol.~34, 2021, pp. 13937--13949.

\bibitem{CaiT}
H.~Touvron, M.~Cord, A.~Sablayrolles, G.~Synnaeve, H.~J\'egou, {Going Deeper With Image Transformers}, in: Proceedings of the IEEE International Conference on Computer Vision (ICCV), 2021, pp. 32--42.

\bibitem{SwinTransformer}
Z.~Liu, Y.~Lin, Y.~Cao, H.~Hu, Y.~Wei, Z.~Zhang, S.~Lin, B.~Guo, {Swin Transformer: Hierarchical Vision Transformer Using Shifted Windows}, in: Proceedings of the IEEE International Conference on Computer Vision (ICCV), 2021, pp. 10012--10022.

\bibitem{LVViT}
Z.-H. Jiang, Q.~Hou, L.~Yuan, D.~Zhou, Y.~Shi, X.~Jin, A.~Wang, J.~Feng, All tokens matter: Token labeling for training better vision transformers, in: Proceedings of the Advances in Neural Information Processing Systems (NeurIPS), Vol.~34, 2021, pp. 18590--18602.

\bibitem{DINO}
M.~Caron, H.~Touvron, I.~Misra, H.~J\'egou, J.~Mairal, P.~Bojanowski, A.~Joulin, Emerging properties in self-supervised vision transformers, in: Proceedings of the IEEE International Conference on Computer Vision (ICCV), 2021, pp. 9650--9660.

\bibitem{XCiT}
A.~Ali, H.~Touvron, M.~Caron, P.~Bojanowski, M.~Douze, A.~Joulin, I.~Laptev, N.~Neverova, G.~Synnaeve, J.~Verbeek, H.~Jegou, Xcit: Cross-covariance image transformers, in: Proceedings of the Advances in Neural Information Processing Systems (NeurIPS), Vol.~34, 2021, pp. 20014--20027.

\bibitem{EsViT}
C.~Li, J.~Yang, P.~Zhang, M.~Gao, B.~Xiao, X.~Dai, L.~Yuan, J.~Gao, Efficient self-supervised vision transformers for representation learning, in: Proceedings of the International Conference on Learning Representations (ICLR), 2022.

\bibitem{CLEViT}
X.~Yu, J.~Wang, Y.~Gao, Cle-vit: Contrastive learning encoded transformer for ultra-fine-grained visual categorization, in: Proceedings of International Joint Conference on Artificial Intelligence (IJCAI), 2023, pp. 4531--4539.

\bibitem{ScopeViT}
X.~Nie, H.~Jin, Y.~Yan, X.~Chen, Z.~Zhu, D.~Qi, Scopevit: Scale-aware vision transformer, Pattern Recognition 153 (2024) 110470.

\bibitem{GradCAM}
R.~R. Selvaraju, M.~Cogswell, A.~Das, R.~Vedantam, D.~Parikh, D.~Batra, Grad-cam: Visual explanations from deep networks via gradient-based localization, in: Proceedings of the IEEE International Conference on Computer Vision (ICCV), 2017.

\bibitem{GradCAM++}
A.~Chattopadhay, A.~Sarkar, P.~Howlader, V.~N. Balasubramanian, Grad-cam++: Visual explanations from deep networks via gradient-based localization, in: Proceedings of Winter Conference on Application of Computer Vision (WACV), 2018, pp. 839--847.

\bibitem{XGrad-CAM}
R.~Fu, Q.~Hu, X.~Dong, Y.~Guo, Y.~Gao, B.~Li, Axiom-based grad-cam: Towards accurate visualization and explanation of cnns, in: Proceedings of the British Machine Vision Conference (BMVC), 2020.

\bibitem{LayerCAM}
P.-T. Jiang, C.-B. Zhang, Q.~Hou, M.-M. Cheng, Y.~Wei, Layercam: Exploring hierarchical class activation maps for localization, {IEEE} Transaction Image Processing 30 (2021) 5875--5888.

\bibitem{AblationCAM}
S.~Desai, H.~G. Ramaswamy, Ablation-cam: Visual explanations for deep convolutional network via gradient-free localization, in: Proceedings of Winter Conference on Application of Computer Vision (WACV), 2020.

\bibitem{ScoreCAM}
H.~Wang, Z.~Wang, M.~Du, F.~Yang, Z.~Zhang, S.~Ding, P.~Mardziel, X.~Hu, Score-cam: Score-weighted visual explanations for convolutional neural networks, in: Proceedings of the IEEE Conference on Computer Vision and Patttern Recognition Workshops (CVPRW), 2020.

\bibitem{TransformerLRP}
H.~Chefer, S.~Gur, L.~Wolf, Transformer interpretability beyond attention visualization, in: Proceedings of the IEEE Conference on Computer Vision and Pattern Recognition (CVPR), 2021, pp. 782--791.

\bibitem{GAE}
H.~Chefer, S.~Gur, L.~Wolf, Generic attention-model explainability for interpreting bi-modal and encoder-decoder transformers, in: Proceedings of the IEEE International Conference on Computer Vision (ICCV), 2021, pp. 397--406.

\bibitem{IIA}
O.~Barkan, Y.~Asher, A.~Eshel, N.~Koenigstein, et~al., Visual explanations via iterated integrated attributions, in: Proceedings of the IEEE International Conference on Computer Vision (ICCV), 2023, pp. 2073--2084.

\bibitem{TaylorDecomposition}
G.~Montavon, S.~Lapuschkin, A.~Binder, W.~Samek, K.-R. M{\"u}ller, Explaining nonlinear classification decisions with deep taylor decomposition, Pattern Recognition 65 (2017) 211--222.

\bibitem{pseudo}
Z.~Pan, W.~Zhang, X.~Yu, M.~Zhang, Y.~Gao, Pseudo-set frequency refinement architecture for fine-grained few-shot class-incremental learning, Pattern Recognition (2024) 110686.

\bibitem{NPID}
Z.~Wu, Y.~Xiong, S.~X. Yu, D.~Lin, Unsupervised feature learning via non-parametric instance discrimination, in: Proceedings of the IEEE Conference on Computer Vision and Pattern Recognition (CVPR), 2018.

\bibitem{MixViT}
X.~Yu, J.~Wang, Y.~Zhao, Y.~Gao, Mix-vit: Mixing attentive vision transformer for ultra-fine-grained visual categorization, Pattern Recognition 135 (2023) 109131.

\bibitem{ReViT}
A.~Diko, D.~Avola, M.~Cascio, L.~Cinque, Revit: Enhancing vision transformers feature diversity with attention residual connections, Pattern Recognition (2024) 110853.

\bibitem{gashisTransformer}
H.~Chen, C.~Li, G.~Wang, X.~Li, M.~M. Rahaman, H.~Sun, W.~Hu, Y.~Li, W.~Liu, C.~Sun, Gashis-transformer: A multi-scale visual transformer approach for gastric histopathological image detection, Pattern Recognition 130 (2022) 108827.

\bibitem{RISE}
V.~Petsiuk, A.~Das, K.~Saenko, {RISE: Randomized Input Sampling for Explanation of Black-box Models}, in: Proceedings of the British Machine Vision Conference (BMVC), 2018.

\bibitem{MeaningfulPerburbation}
R.~C. Fong, A.~Vedaldi, Interpretable explanations of black boxes by meaningful perturbation, in: Proceedings of the IEEE International Conference on Computer Vision (ICCV), 2017, pp. 3429--3437.

\bibitem{ResNet}
K.~He, X.~Zhang, S.~Ren, J.~Sun, {Deep Residual Learning for Image Recognition}, in: Proceedings of the IEEE Conference on Computer Vision and Pattern Recognition (CVPR), 2016.

\bibitem{ReLU}
X.~Glorot, A.~Bordes, Y.~Bengio, {Deep Sparse Rectifier Neural Networks}, in: Proceedings of International Conference on Artificial Intelligence and Statistic (AISTATS), Vol.~15, 2011, pp. 315--323.

\bibitem{ImageNet2012}
O.~Russakovsky, J.~Deng, H.~Su, J.~Krause, S.~Satheesh, S.~Ma, Z.~Huang, A.~Karpathy, A.~Khosla, M.~Bernstein, et~al., Imagenet large scale visual recognition challenge, International Journal of Computer Vision 115 (2015) 211--252.

\bibitem{SanityCheck}
J.~Adebayo, J.~Gilmer, M.~Muelly, I.~Goodfellow, M.~Hardt, B.~Kim, Sanity checks for saliency maps, in: Proceedings of the Advances in Neural Information Processing Systems (NeurIPS), 2018, pp. 9525--9536.

\bibitem{spare}
X.~Yu, Y.~Zhao, Y.~Gao, Spare: Self-supervised part erasing for ultra-fine-grained visual categorization, Pattern Recognition 128 (2022) 108691.

\end{thebibliography}
\newpage
\end{document}